\documentclass[letterpaper, 10 pt, conference]{ieeeconf}  
\IEEEoverridecommandlockouts                              
\usepackage{amsmath}
\usepackage[table,xcdraw]{xcolor}
\usepackage{amssymb}
\usepackage{graphicx}
\usepackage[hidelinks]{hyperref}
\usepackage{caption}
\usepackage{float}
\usepackage{moresize}
\usepackage[ruled,noend]{algorithm2e}
\usepackage{booktabs}
\usepackage{comment}
\usepackage{subfigure}
\usepackage{afterpage}

\newcommand{\method}[1]{\textit{#1}}

\usepackage{etoolbox}
\makeatletter
\patchcmd{\@algocf@start}
  {-1.5em}
  {0pt}
  {}{}
\makeatother

\title{\LARGE \bf
Dream2Real: Zero-Shot 3D Object Rearrangement\\with Vision-Language Models
}

\author{Ivan Kapelyukh$^{*1,2}$, Yifei Ren$^{*1}$, Ignacio Alzugaray$^{2}$, Edward Johns$^{1}$
\thanks{$^*$ Joint first authorship. $^{1}$ The Robot Learning Lab at Imperial College London. $^2$ The Dyson Robotics Lab at Imperial College London.}
}

\begin{document}

\twocolumn[{
\renewcommand\twocolumn[1][]{#1}
\maketitle
\begin{center}
    \centering
    \captionsetup{type=figure}
    \includegraphics[width=0.7\linewidth]{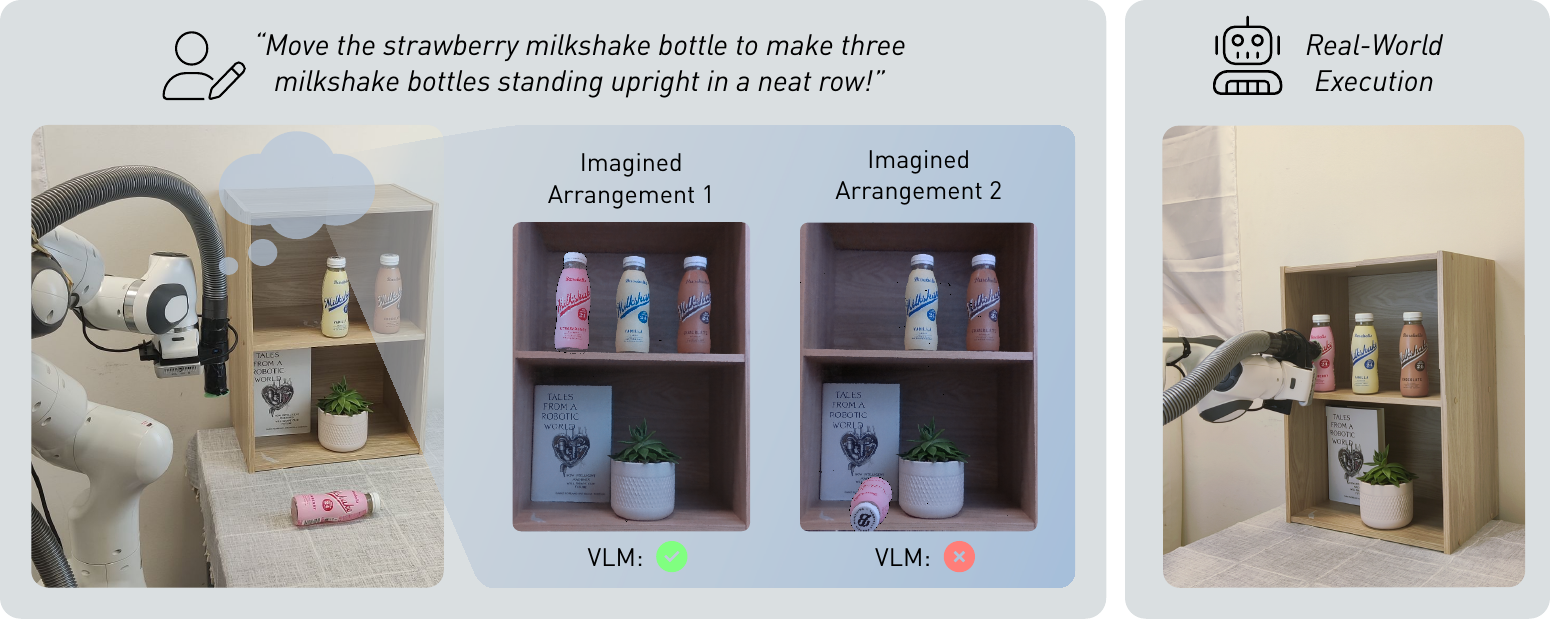}
    \caption{Dream2Real enables a robot to imagine, and then evaluate, virtual rearrangements of scenes. First, the robot builds an object-centric NeRF of a scene. Then, numerous reconfigurations of the scene are rendered as 2D images. Finally, a VLM evaluates these according to the user instruction, and the best is then physically created using pick-and-place.}
    \label{fig:fig1}
\end{center}
}]

\thispagestyle{empty}
\pagestyle{empty}

\begin{abstract}
We introduce Dream2Real, a robotics framework which integrates vision-language models (VLMs) trained on 2D data into a 3D object rearrangement pipeline. This is achieved by the robot autonomously constructing a 3D representation of the scene, where objects can be rearranged virtually and an image of the resulting arrangement rendered. These renders are evaluated by a VLM, so that the arrangement which best satisfies the user instruction is selected and recreated in the real world with pick-and-place. This enables language-conditioned rearrangement to be performed zero-shot, without needing to collect a training dataset of example arrangements. Results on a series of real-world tasks show that this framework is robust to distractors, controllable by language, capable of understanding complex multi-object relations, and readily applicable to both tabletop and 6-DoF rearrangement tasks. Videos are available on our webpage at: \textcolor{blue}{\href{https://www.robot-learning.uk/dream2real}{https://www.robot-learning.uk/dream2real}.}
\vspace{-0.2cm}
\end{abstract}

\let\thefootnote\relax\footnotetext{$^*$ Joint first authorship. $^{1}$ The Robot Learning Lab at Imperial College London. $^2$ The Dyson Robotics Lab at Imperial College London.}
\section{Introduction}

Consider being asked to perform a task such as arranging bottles in a row, as in Figure \ref{fig:fig1}. To achieve this, humans might first imagine the goal state that should be created according to the instructions. This imagined arrangement should be physically valid, visually natural (e.g. the bottles are not upside down), and semantically correct for the given task. In this paper, we study how robots can imagine (or \textit{dream}) new configurations of scenes, and then evaluate them to select a suitable goal state. This leads to our language-conditioned 6-DoF object rearrangement framework, Dream2Real.

Designing Dream2Real opens up two key questions. Firstly, \textit{How can a robot imagine a new scene configuration?}, and secondly, \textit{How can a robot evaluate an imagined configuration according to a language command?}

We begin with the second question, and note that vision-language models (VLMs) have shown great promise in enabling robots to understand how language instructions relate to the scene before them \cite{cliport,rt2}, enabling generalisation across many objects and tasks. One such VLM is CLIP \cite{clip}. By training on hundreds of millions of captioned images from the Web (including images of object arrangements), CLIP learns to predict how closely an image matches a text description. This is exactly the reasoning a robot requires when evaluating novel scene arrangements it has imagined with respect to a user's language instruction.

For the first question, we note that VLMs typically operate on 2D images, and yet robots operate in 3D spaces. Therefore, we propose an approach where the robot autonomously builds a 3D representation of the scene using two Neural Radiance Fields (NeRF) \cite{mildenhall2021nerf}, one for the object to be moved and one for the background. This representation can be arranged into new ``imaginary'' configurations, and then rendered to yield a photorealistic 2D image for the VLM to evaluate.

We address the difficult technical challenges of first autonomously constructing a 3D representation of the scene which can be rearranged in imagination, and secondly interfacing this effectively with 2D VLMs. As well as proposing the overall framework, we make several technical contributions when we instantiate this framework in a practical, real-world rearrangement system, including: automatically identifying distractor objects using a language model and hiding them from the VLM, introducing normalising captions (a novel technique) which encourages the VLM to focus on spatial relations, identifying objects in the scene more reliably by aggregating information across views, and constructing collision meshes which avoids the need to render and evaluate physically impossible configurations.

Through a series of real-world experiments, we share intriguing findings about the applicability of VLMs to rearrangement. We show that our novel approach of imagining and evaluating arrangements outperforms recent work on VLMs for tabletop rearrangement \cite{dallebot}. By conducting a range of ablation studies, we evaluate which techniques are most useful and in which situations. We find that our approach is robust to distractors, can evaluate complex many-object spatial relations, and is readily applicable to 3D scenes.

Integrating a VLM and a NeRF-based representation with editable poses in this novel way yields several strengths. First, Dream2Real is \textbf{zero-shot}, as it applies VLMs to object rearrangement without requiring a training dataset of example arrangements to be collected. Second, it achieves full \textbf{6-DoF rearrangement}, whereas prior zero-shot work \cite{dallebot} is limited to top-down scenes. Third, we show that the use of VLMs for \textbf{visual evaluation} of imagined goal states is better than asking VLMs to predict the goal state directly.

To the best of our knowledge, this is the first method which performs 6-DoF rearrangement zero-shot by using the web-scale visual reasoning of VLMs.

\section{Related Work}

\textbf{Predicting goal poses} is a key challenge in object rearrangement \cite{rearrangement}. Prior work classifies the correct receptacle in which to place an object \cite{organisational-principles-classification,tidee,consor}, or predicts a continuous goal pose \cite{distilling-vlm,efficient-interpretable,sg-bot}. Another approach learns rearrangement from full demonstrations \cite{transporters,cliport}. The prediction of goal poses can be conditioned in several ways. Some methods allow users to specify relational predicates \cite{semantic-stable,ebm-rearrange}. Others learn a personalised representation of user preferences \cite{abdo-shelves,neatnet,adaptable-planners}. User instructions may also be expressed in free-form language. StructFormer \cite{structformer} trains a language-conditioned transformer on a synthetic dataset of over 100,000 rearrangement sequences. To better avoid collisions, StructDiffusion \cite{structdiffusion} learns a distribution over desirable poses using a diffusion model. SceneScore \cite{scenescore} uses an energy-based model, thus learning to evaluate whether a given arrangement is desirable. These methods typically require training on thousands of example arrangements \cite{structformer,cabinet,rpdiff}. This is effective for specific tasks, but is difficult to scale to unstructured environments such as homes, because of the difficulty of generating realistic training data.

Instead, \textbf{large language models} (LLMs) pre-trained on web-scale data can be used to predict arrangements \cite{housekeep,llm-rearrange,tidybot}. These are effective for high-level planning, but language models typically lack the visual perception needed to e.g. assess whether an object is oriented correctly. Our framework uses VLMs to \textit{visually} evaluate imagined scenes. The closest work to ours is DALL-E-Bot \cite{dallebot}, which achieves visual rearrangement zero-shot by using a web-scale diffusion model to generate a goal image, and then matching that to the real scene. However, this is limited to top-down scenes. Experiments show that our evaluative approach is more robust than predicting a goal state directly.

Beyond rearrangement, \textbf{vision-language models} and LLMs have proven to be useful for bringing web-scale semantic understanding to embodied agents \cite{rt2,saycan,moo,rosie,cacti,genaug,voxposer}. VLMs in particular can also be used to connect user instructions in natural language with the robot's visual perception \cite{cliport,distilled}. Closer to our work, VLMs and LLMs can also act as reward signals to train robot policies \cite{llm-reward,liv,zest}. In this work, we show how the web-scale visual prior of VLMs can solve 3D rearrangement tasks zero-shot without any further policy training.

\textbf{3D reconstruction} research continues to yield useful techniques for robotics \cite{distilled,fit-ngp,lerf-togo,mira,driess-compnerf}. Implicit neural representations such as NeRF \cite{mildenhall2021nerf} have shown a strong ability to produce photorealistic renders from novel views. Instant-NGP \cite{muller2022instant} greatly accelerates the training and rendering speed of NeRF using multiresolution hash encoding, enabling real-time rendering. Scenes can also be decomposed with object-level reconstruction, as shown in works like vMAP \cite{kong2023vmap}. Related to our method, some works \cite{wang2022clip, mirzaei2022laterf,semantic-abstraction,conceptfusion,clip-fields} also combine 3D representations with LLMs, but do not focus on zero-shot 3D robotic rearrangement.

\section{Method}

\begin{figure*}[ht]
    \centerline{\includegraphics[width=1.0\linewidth]{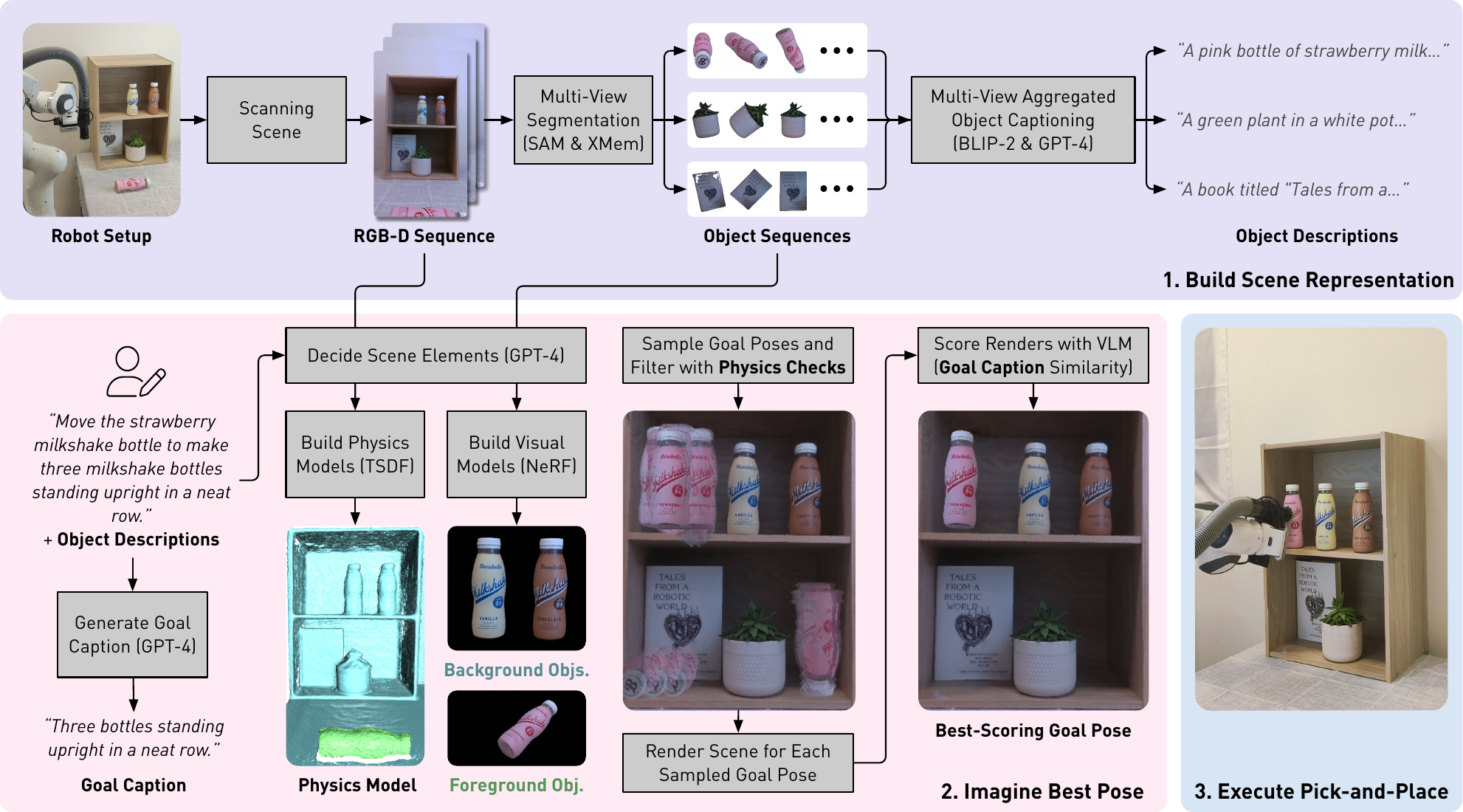}}
    \caption{The Dream2Real pipeline. The robot first autonomously builds a model of the scene. Then the user instruction is used to determine which object should be moved, and so the robot can imagine new configurations of the scene and score them using a VLM. Finally, the highest-scoring pose is used as the goal for pick-and-place to complete the rearrangement.}
    \label{fig:pipeline}\vspace{-0.25cm}
\end{figure*}

The core problem we address is determining a goal pose for an object given a language instruction, so that a robot can pick-and-place it into that goal pose. Our modular pipeline shown in Figure \ref{fig:pipeline}. First, the robot builds a scene representation (Section \ref{s:observing}). Then it processes the user instruction to understand the task (Section \ref{s:interpreting}). Next, task-specific models are constructed for foreground and background objects (Section \ref{s:task-specific}). By sampling and evaluating arrangements based on these models, the best goal pose for the object to be moved can be determined (Section \ref{s:dream}). Finally, the robot executes the rearrangement (Section \ref{s:execution}). Further implementation and evaluation details are provided in the supplementary material on our project webpage.

\subsection{Observing the Scene}\label{s:observing}
In our framework, the robot constructs as much of its scene representation as possible before a user instruction is received (i.e. when a robot first observes a scene), thus reducing the instruction execution delay. Therefore, the robot starts by autonomously scanning the scene to collect a set of RGB-D images which will be used later for building the object-level visual and physics models. In our experiments, we use a hemispherical camera trajectory facing the scene centre.
We segment the scene into objects by running Segment Anything (SAM) \cite{kirillov2023segment} on the first frame from our trajectory where all objects are assumed to be at least partially visible.
Those objects are tracked across the other views using XMem \cite{cheng2022xmem}, handling the data association problem across frames. Given the tracked objects, we extract image crops from each of the views in which they are visible and apply BLIP-2 \cite{blip2} to retrieve a per-crop caption.
Since captions for the same object may differ across views, we use an LLM (GPT-4) \cite{gpt4} to aggregate these into one coherent object description. We find that this multi-view aggregation allows objects to be identified more reliably.
An example object description produced by the language model is: \textit{``A pink bottle of strawberry milk or juice with a red label, white cap, and barcode on it, sitting on a table or white surface.''}

\subsection{Interpreting User Instructions}\label{s:interpreting}

Once the user instruction is received, the robot must process it to understand the task. We automatically extract four key items from the user instruction using a language model (GPT-4): the \textit{movable object}, \textit{relevant objects}, the \textit{goal caption} and the \textit{normalising caption}. E.g. suppose that the user instruction is \textit{``put the apple inside the bowl''}. The movable object here is the apple, since it is the one which should be moved to fulfil the instruction. Relevant objects are those which the VLM should observe to evaluate whether the user instruction is fulfilled (apple and bowl). This technique avoids showing distractor (irrelevant) objects to CLIP, which we show is crucial for performance. The goal
caption is a description of the desired final state after the
instruction has been fulfilled. In this example, it would be: \textit{``an apple inside a bowl''}. Lastly, the normalising caption is a description of the scene that remains neutral to the pose of the object being moved. Typically, GPT-4 simply returns a list of objects within the scene, e.g. \textit{``an apple and a bowl''}. This will be used later for normalising CLIP scores (Section \ref{s:dream}).

\subsection{Building Task-Specific Visual \& Physics Models}\label{s:task-specific}
We now describe how to construct the physics models which we use to check imagined arrangements for collisions, and the visual models that we use for rendering those arrangements. We separate the scene into the foreground (the movable object) and the background (relevant objects excluding the movable object), then build two separate visual models accordingly. We use NeRF (specifically Instant-NGP \cite{muller2022instant}) because of its high visual realism and speed for both training and rendering. In detail, for both foreground and background objects, using masks from XMem, we assign pixels outside of the corresponding masks 0 alpha value. During NeRF training, this encourages the space around the object to be represented as empty, which will later allow us to freely move this object around the scene and render it from novel poses. Since we move the entire foreground NeRF, this empty space supervision is important to allow the two NeRFs to be rendered together correctly.
To build the physics models, we combine depth images from across views to create a separate foreground and background Truncated Signed Distance Function (TSDF) \cite{curless1996volumetric, Zhou2018}, which we find achieves more accurate geometry than extracting a mesh from Instant-NGP.

\subsection{Dreaming the Best Pose}\label{s:dream}
Now that we have a separate, movable model for the foreground object, we can sample many different poses for it and evaluate each of these ``imaginary'' arrangements, to find a desirable pose. We find experimentally that a straightforward sampling strategy where we sample positions in a dense, regular 3D grid covering the scene (and sample orientations from discretised bins) works well. We move (virtually) the movable object's physics model to each of the sampled poses in turn and check for physical validity, i.e. the object must not be in collision with the scene or unsupported in free space. Thus we avoid rendering and evaluating invalid poses. Then, for each valid pose, we render the foreground NeRF as if the object were in that pose, and combine this with the background NeRF render (using a similar approach to \cite{kong2023vmap}). We render the NeRFs from a fixed camera pose from our scanning trajectory facing the centre of the scene.

We now have a rendered RGB image for each sampled goal state, which can be evaluated with a web-scale VLM. We batch-compute the CLIP similarity \cite{clip} between the image of each arrangement and the goal caption. We also divide this similarity score by the similarity of the image with the normalising caption. Intuitively, we want the overall similarity score to focus only on whether the spatial relation requested by the user is satisfied or not in the image. We show experimentally that this is important for performance. We also implement \textit{spatial smoothing}: a Gaussian smoothing filter is applied on the 3D grid of scores to reduce the score of outlier poses, which have a high score but are surrounded by many low-scoring poses. Finally, we select the highest-scoring sampled pose as the goal pose for the movable object.

\subsection{Robot Execution}\label{s:execution}
Once the goal pose has been determined, the robot executes the rearrangement using pick-and-place. For our grasping module we use the FC-GQCNN from DexNet 4.0 \cite{dexnet}, but any off-the-shelf grasping method can easily be applied \cite{reorientbot,diffusion-grasp}. We then use inverse kinematics and a motion planner (RRT-Connect \cite{rrt-connect}) to find a path between the pick and place poses which avoids collisions, using the object collision meshes that the robot constructed previously.

\section{Experiments}
In our experiments, we investigate three key research questions: (1) Can Dream2Real predict goal poses zero-shot based on language instructions in multi-task scenes (Section \ref{ss:shopping})? (2) Does this method understand many-object relations (Section \ref{ss:pool})? (3) Can this approach make 2D VLMs work with 3D scenes? (Section \ref{ss:shelf})? We also demonstrate the physical execution of rearrangement in Section \ref{ss:exec-exps}.

\textbf{Hardware setup.} For real-world evaluation we use a 7-DoF Franka Panda with a compliant suction gripper and a wrist-mounted Intel RealSense D435i RGB-D camera.

\begin{figure}[btp]
    \centerline{\includegraphics[width=1.0\linewidth]{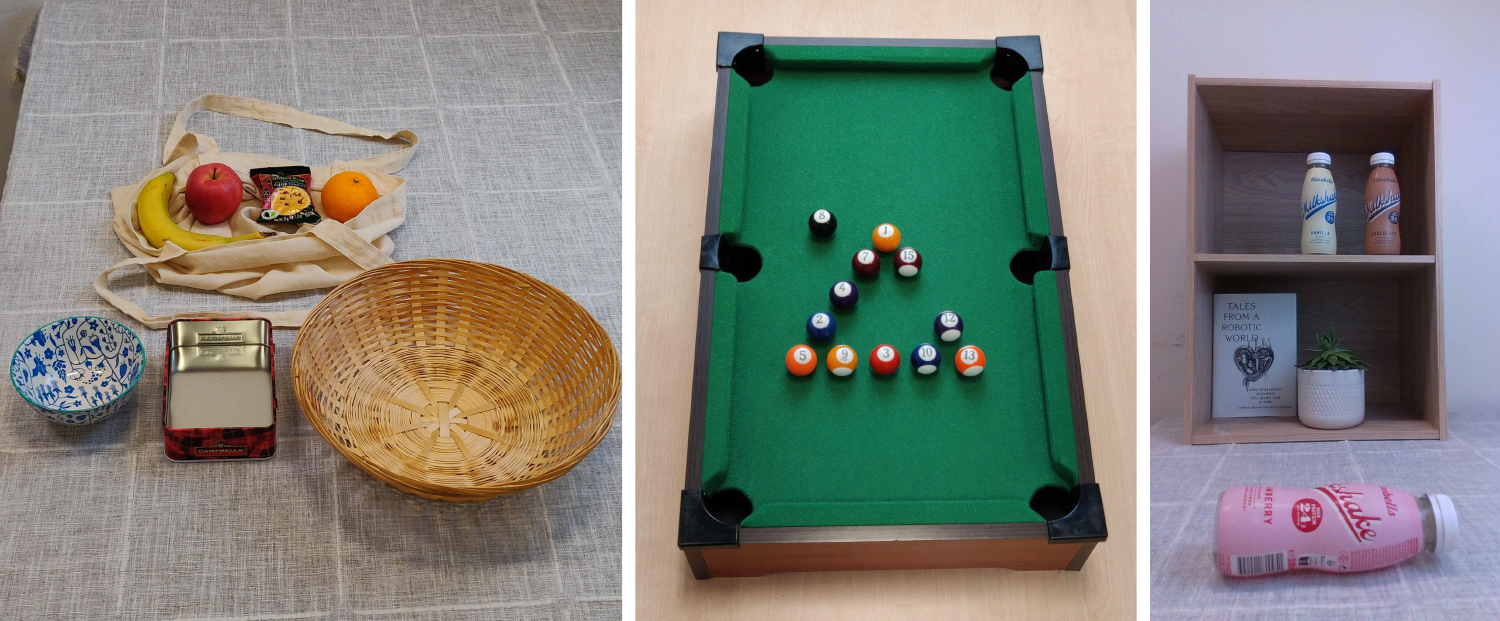}}
    \caption{The shopping, pool ball, and shelf scenes.}
    \label{fig:all-scenes}
    \vspace{-0.3cm}
\end{figure}

\textbf{Scenes and tasks}. We evaluate on 10 rearrangement tasks across 3 distinct real-world scenes, as shown in Figure \ref{fig:all-scenes}. These are described in detail in the following sections.

\textbf{Evaluation metric}. As the primary contribution of this paper is a method for determining a goal pose, our evaluation focuses on whether the predicted goal pose is correct. We measure task success using success regions. This allows us to efficiently and fairly evaluate many variations of our method, by controlling for noise that would arise from physical execution. Physical execution is evaluated as part of the whole pipeline in Section \ref{ss:exec-exps}.

\textbf{Baselines.} (1) Since our method is zero-shot, it cannot be compared fairly against methods which require thousands of example arrangements to be collected \cite{structformer}. Therefore we compare against \method{DALL-E-Bot} \cite{dallebot}, a method for zero-shot rearrangement with VLMs. It uses a diffusion model to generate a goal image, and is restricted to 2D top-down scenes. (2) We also compare with a variant of our method, \method{D2R-One-View}, which uses only the first camera view throughout the whole pipeline (including object captioning), avoiding the need for data collection. Instead of a NeRF, a colour point cloud is rendered as the visual model. (3) Next, the \method{D2R-Distract} ablation does not use GPT-4 to filter out irrelevant objects (distractors). (4) The \method{Physics-Only} baseline does not use CLIP to evaluate poses: instead, it uses a random physically valid pose. (5) \method{D2R-No-Norm} does not use normalising captions. (6) \method{D2R-Vis-Prior} investigates the visual prior of CLIP: it does not use normalising captions for normalisation. Instead, it uses them as goal captions. E.g. if the goal caption was previously ``an apple inside a bowl'', then it now becomes ``an apple and a bowl''. This tests whether CLIP knows a natural pose for the apple without being told it should go in the bowl. (7) Finally, \method{D2R-No-Smooth} ablates the spatial smoothing technique.

\subsection{Zero-Shot Multi-Task Rearrangement}\label{ss:shopping}
First we evaluate on a scene we refer to as the shopping scene, where many tasks are possible. We choose a top-down scene to allow a comparison against DALL-E-Bot \cite{dallebot}. The 5 instructions (i.e. 5 tasks) for this scene are: (1) \textit{``put the apple inside the blue and white bowl''}, (2) \textit{``put the apple beside the blue and white bowl''}, (3) \textit{``put the cookies inside the square metal box''}, (4) \textit{``put the orange inside the blue and white bowl''}, and (5) \textit{``put the banana inside the wicker basket''}. We sample object positions but not orientations here. We run 7 repeats for each method-task combination, for a total of 280 goal pose predictions (in imagination). In between repeats, we shuffle object positions and re-scan the scene.

\begin{table*}[t]
    \centering
    \caption{Success rates for the shopping scene (\%).}
    \label{tab:results-shopping}
    \begin{tabular}{lcccccc}
        \toprule
        Method & \textit{apple in bowl} & \textit{apple beside bowl} & \textit{orange in bowl} & \textit{cookies in box} & \textit{banana in basket} & \textit{mean} \\
        \midrule
        Physics-Only  & 0 & 57 & 14 & 0 & 14 & 17 \\
        D2R-Distract  & 0 & 71 & 14 & 0 & 0 & 17 \\
        D2R-Vis-Prior  & 0 & 71 & 14 & 0 & 0 & 17 \\
        DALL-E-Bot \cite{dallebot} & 14 & 29 & 0 & \textbf{43} & 86 & 34 \\
        D2R-No-Norm  & 29 & 71 & 71 & 0 & 29 & 40 \\
        D2R-One-View  & 71 & 14 & 57 & 29 & \textbf{100} & 54 \\
        Dream2Real & \textbf{100} & 71 & \textbf{100} & \textbf{43} & \textbf{100} & 83 \\
        D2R-No-Smooth  & \textbf{100} & \textbf{86} & \textbf{100} & \textbf{43} & \textbf{100} & \textbf{86} \\
        \bottomrule
    \end{tabular}
    \vspace{-0.3cm}
\end{table*}

\begin{figure}[btp]
    \centering
    \subfigure{\label{fig: apple in bowl heatmap}
        \includegraphics[width=0.3\linewidth]{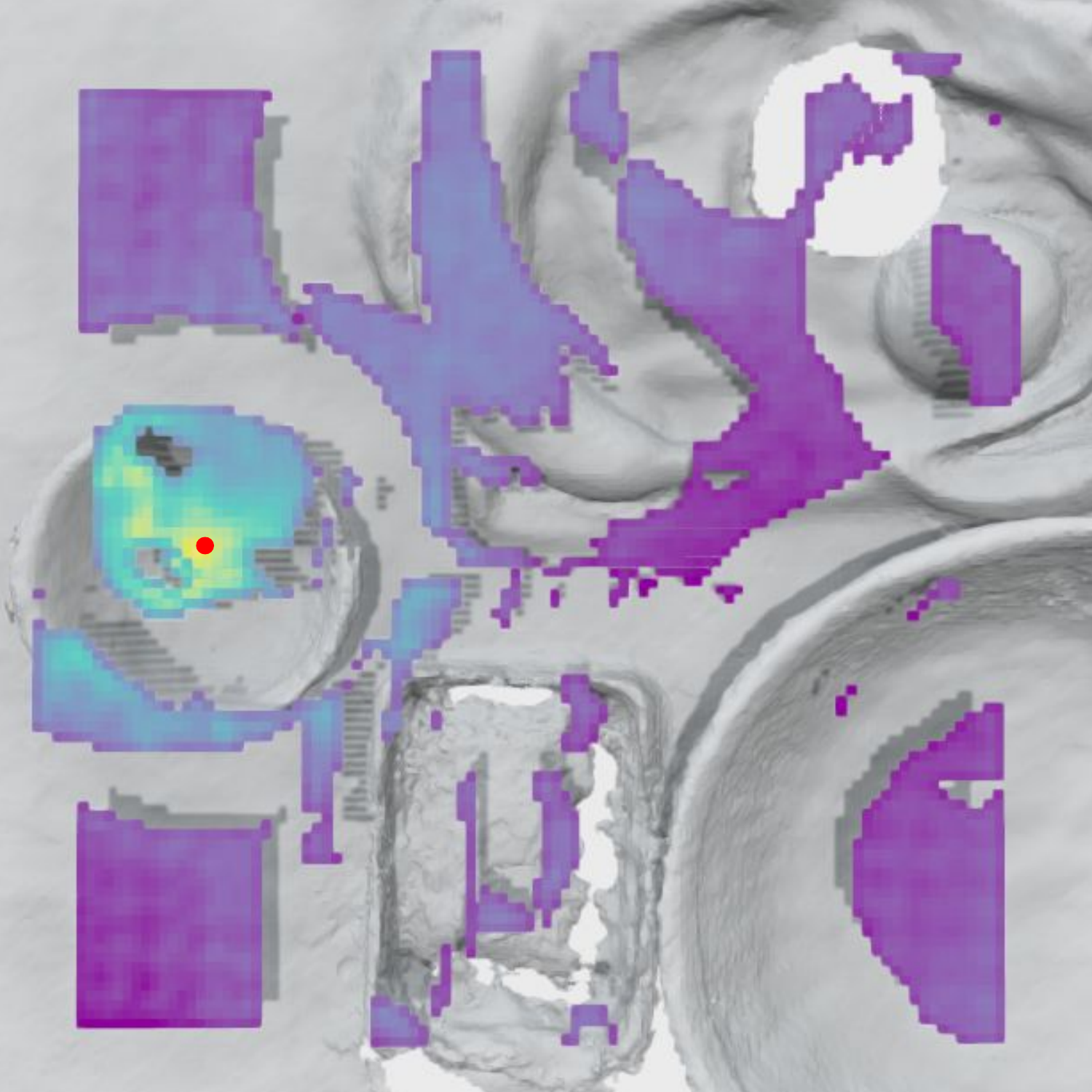}
    }\hspace{-0.22cm}
    \subfigure{
        \label{fig: apple in bowl rendering}
        \includegraphics[width=0.3\linewidth]{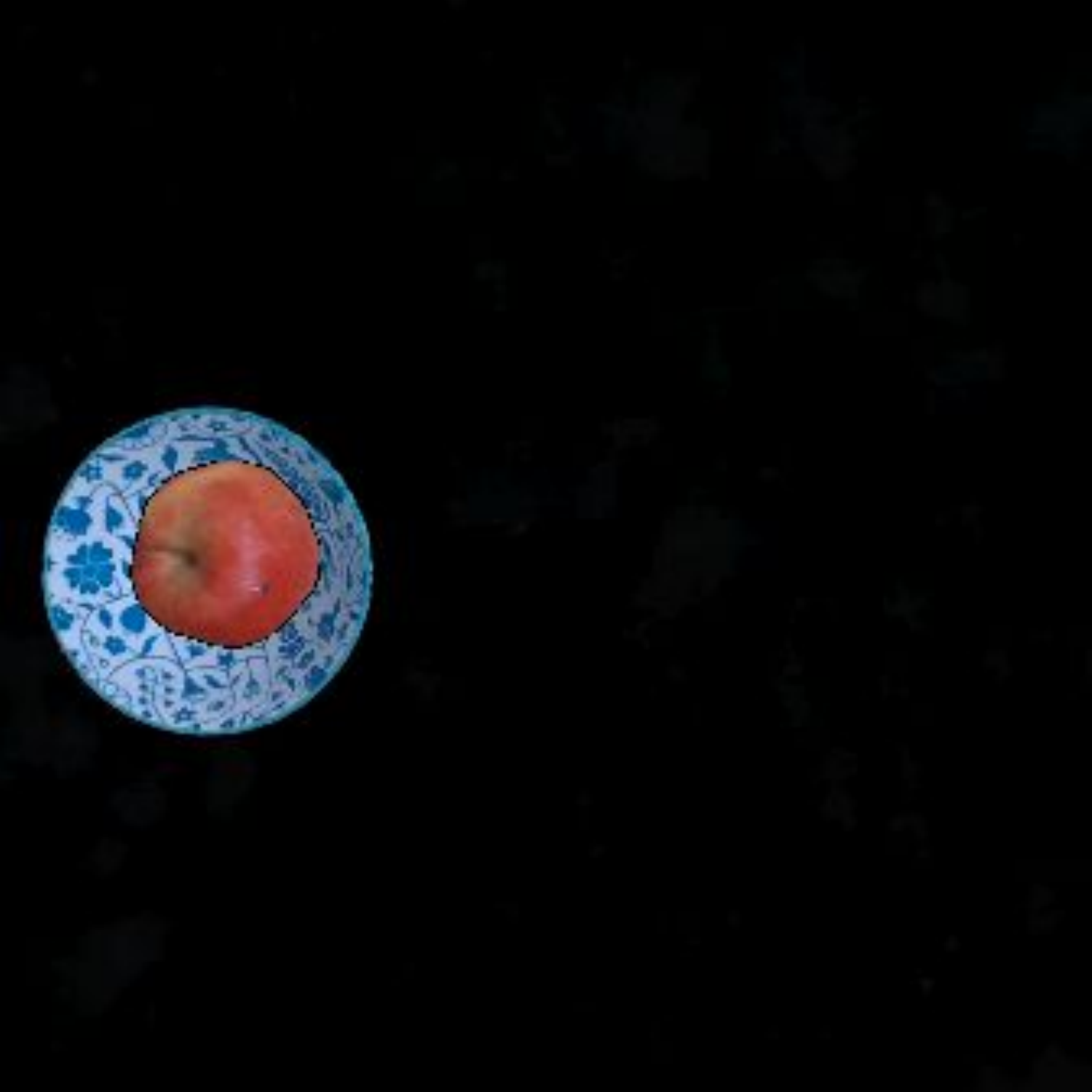}
    }
    \setcounter{subfigure}{0}\\\vspace{-0.22cm}
    \subfigure[CLIP score heatmap]{
        \label{fig: apple beside bowl heatmap}
        \includegraphics[width=0.3\linewidth]{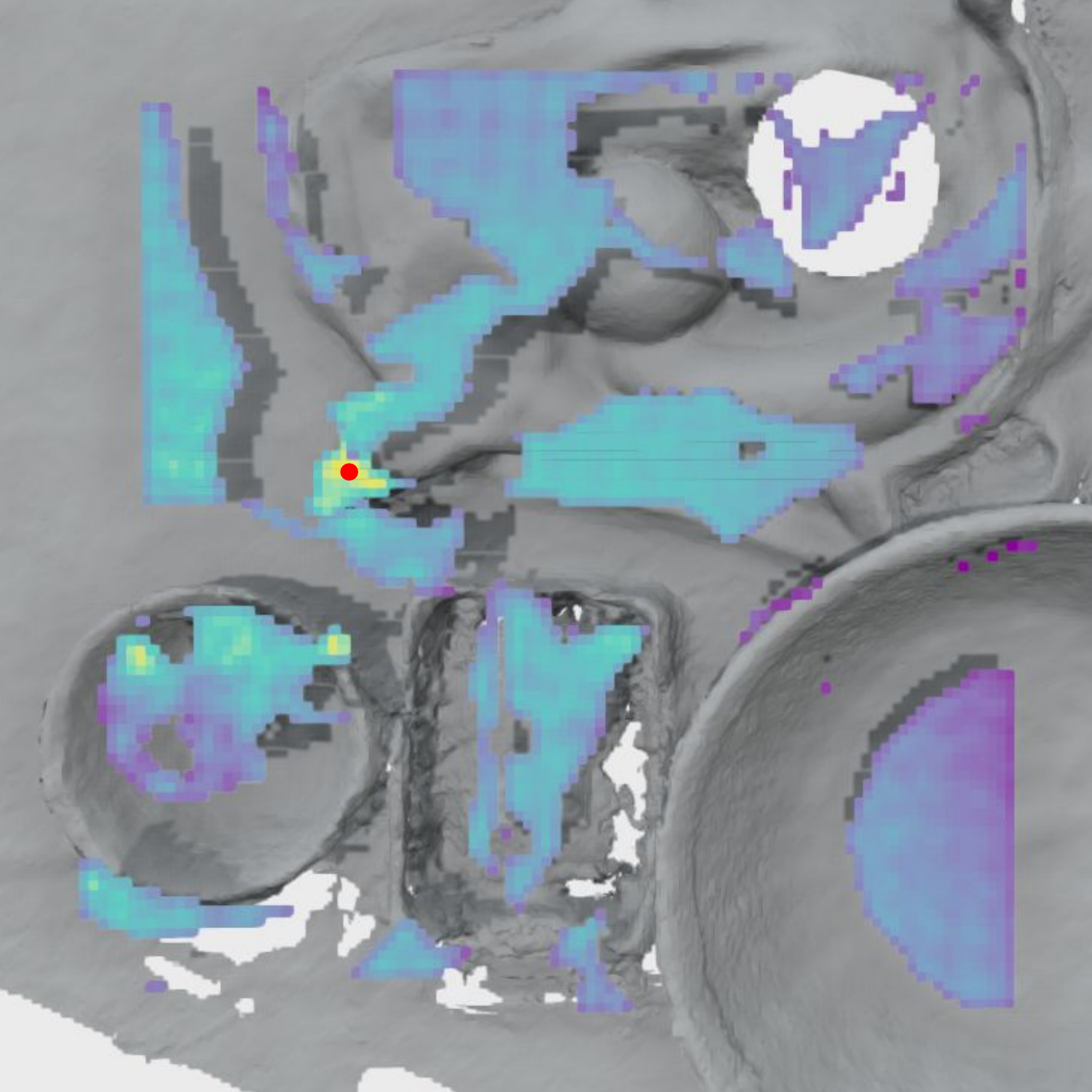}
    }\hspace{-0.22cm}
    \subfigure[Max score render]{
        \label{fig: apple beside bowl rendering}
        \includegraphics[width=0.3\linewidth]{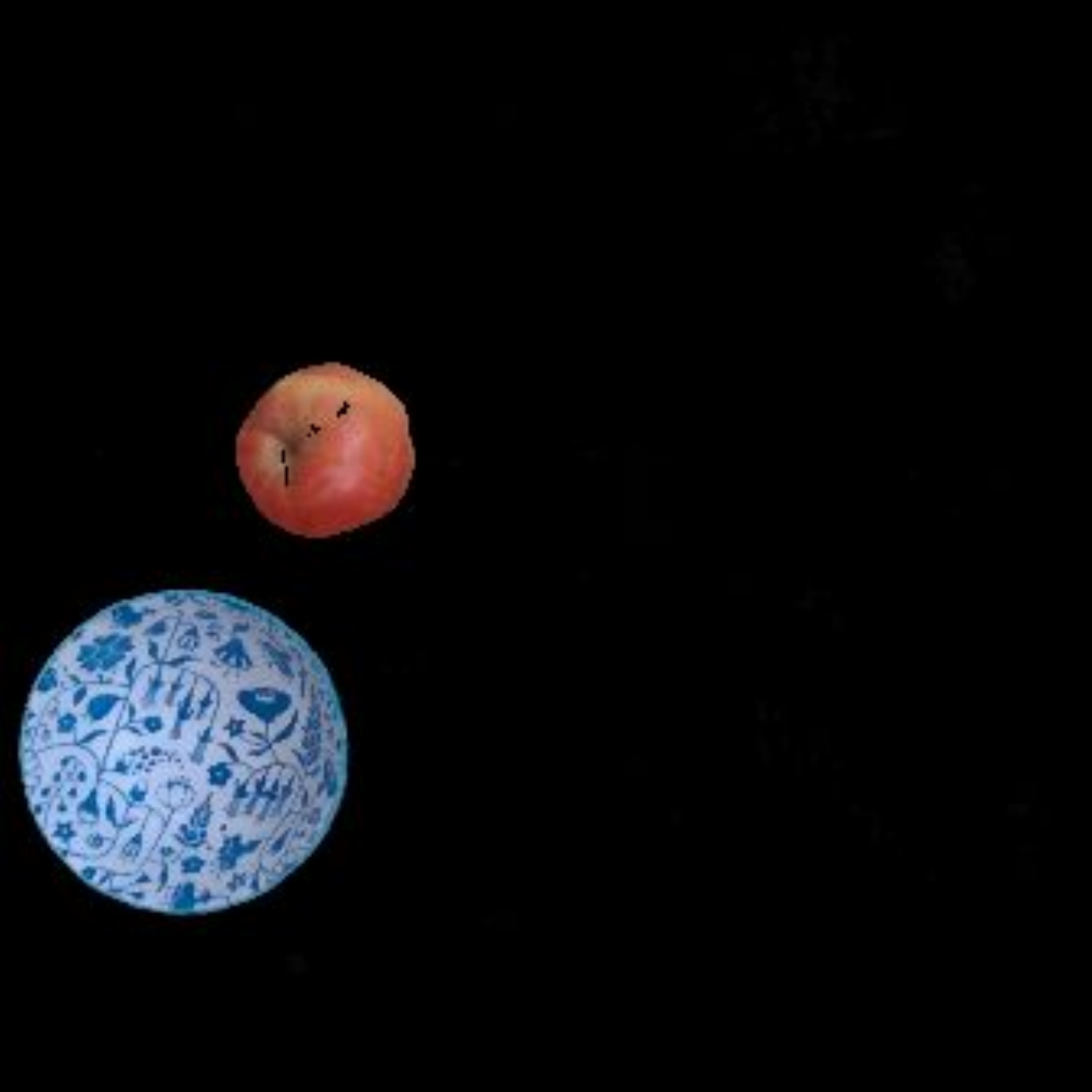}
    }
    \caption{Qualitative results from the shopping scene for the tasks ``apple in bowl'' (top row) and ``apple beside bowl'' (bottom row). Figure \ref{fig:all-scenes} shows the full shopping scene. In the heatmaps (overlaid on the TSDF of the scene), yellow indicates high-scoring positions of the apple, whereas dark blue indicates low-scoring regions, and colliding poses are not included. The red dot highlights the highest-scoring position. The highest-scoring render is shown on the right.}
    \label{fig:shopping-qual}
\end{figure}

Qualitative results are in Figure \ref{fig:shopping-qual}, showing a heatmap of CLIP scores next to the best-scoring render. Table \ref{tab:results-shopping} shows quantitative results. Our method significantly outperforms \method{DALL-E-Bot} \cite{dallebot} (83\% vs 34\% mean success rate). This is due to a key difference in how our approaches use VLMs: DALL-E-Bot is \textit{predictive}, i.e. it generates a goal image and attempts to match those objects to the real world. However, DALL-E-Bot very often generates images with a different number of objects to the real world, and so (despite its filtering techniques) it matches the real object to a generated object in the wrong place. Dream2Real is \textit{evaluative}, using a VLM to score sampled arrangements of the real objects, thus avoiding this difficult matching problem. DALL-E-Bot is also affected by distractors, whereas our method automatically hides them from the VLM. \method{D2R-Vis-Prior}'s lower performance suggests that conditioning the visual prior on language is important. Our method also doubles the success rate of \method{D2R-No-Norm}, showing that normalising captions are effective for these tasks in forcing CLIP to focus on the spatial relations in the instruction. \method{D2R-No-Smooth}'s high performance shows that CLIP works well here even without outlier filtering. The \method{D2R-Distract} ablation shows that our technique of only showing relevant objects to the VLM is crucial for performance on cluttered scenes. This experiment shows that Dream2Real can succeed at everyday rearrangement tasks zero-shot.

\subsection{Multi-Object Geometric Relations}\label{ss:pool}

\begin{figure}[tbp]
    \centering
    \subfigure{\includegraphics[width=0.3\linewidth]{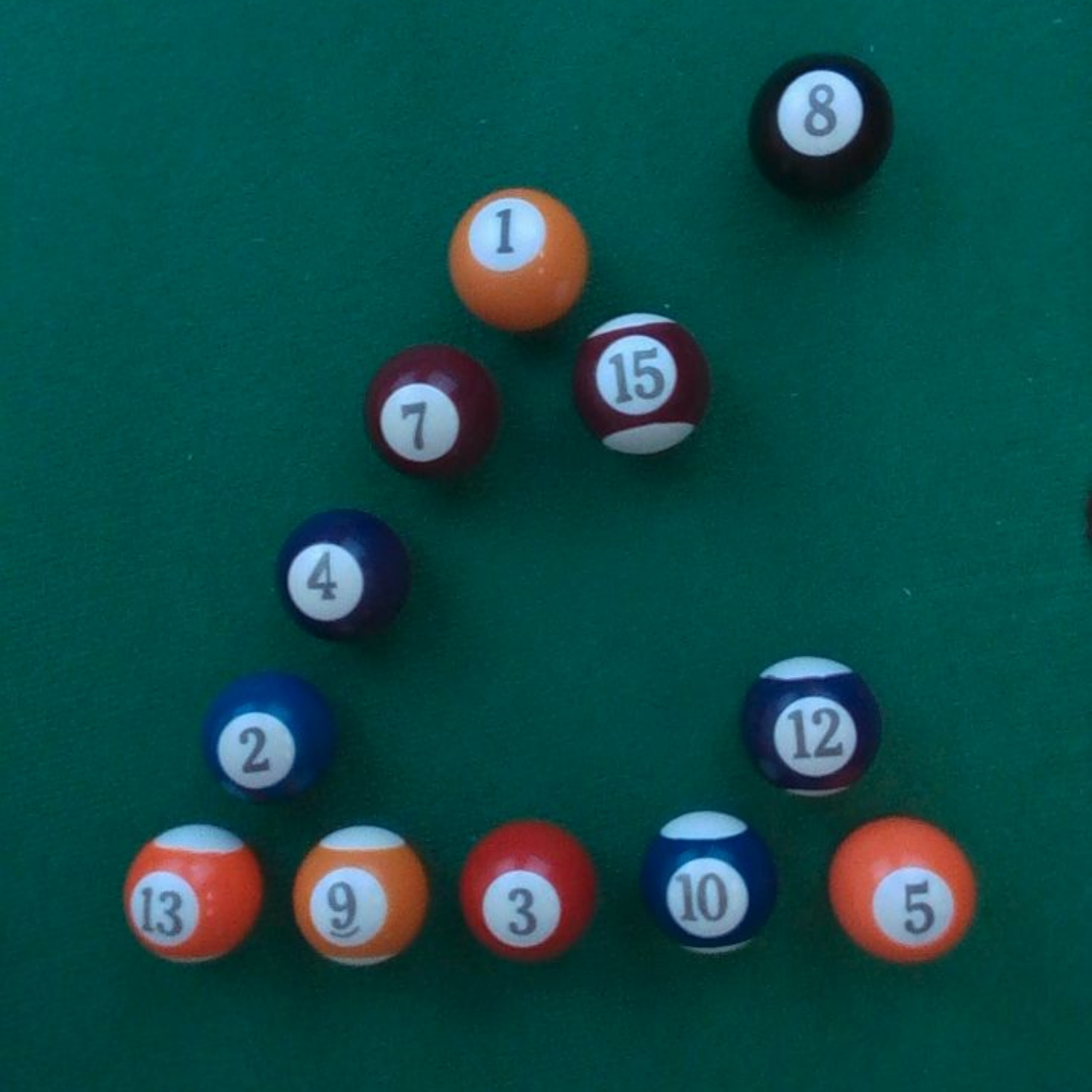}}
    \subfigure{\includegraphics[width=0.3\linewidth]{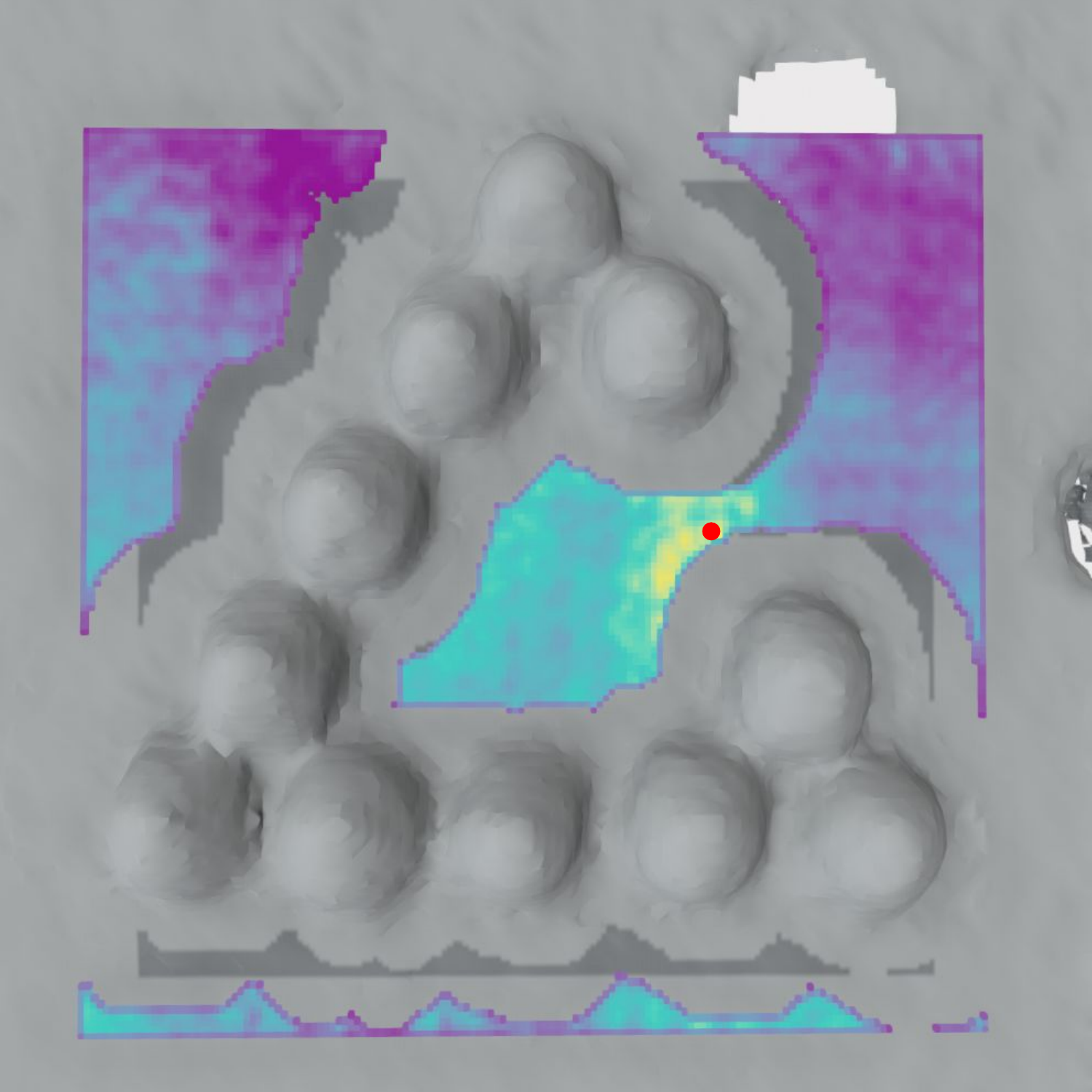}}
    \subfigure{\includegraphics[width=0.3\linewidth]{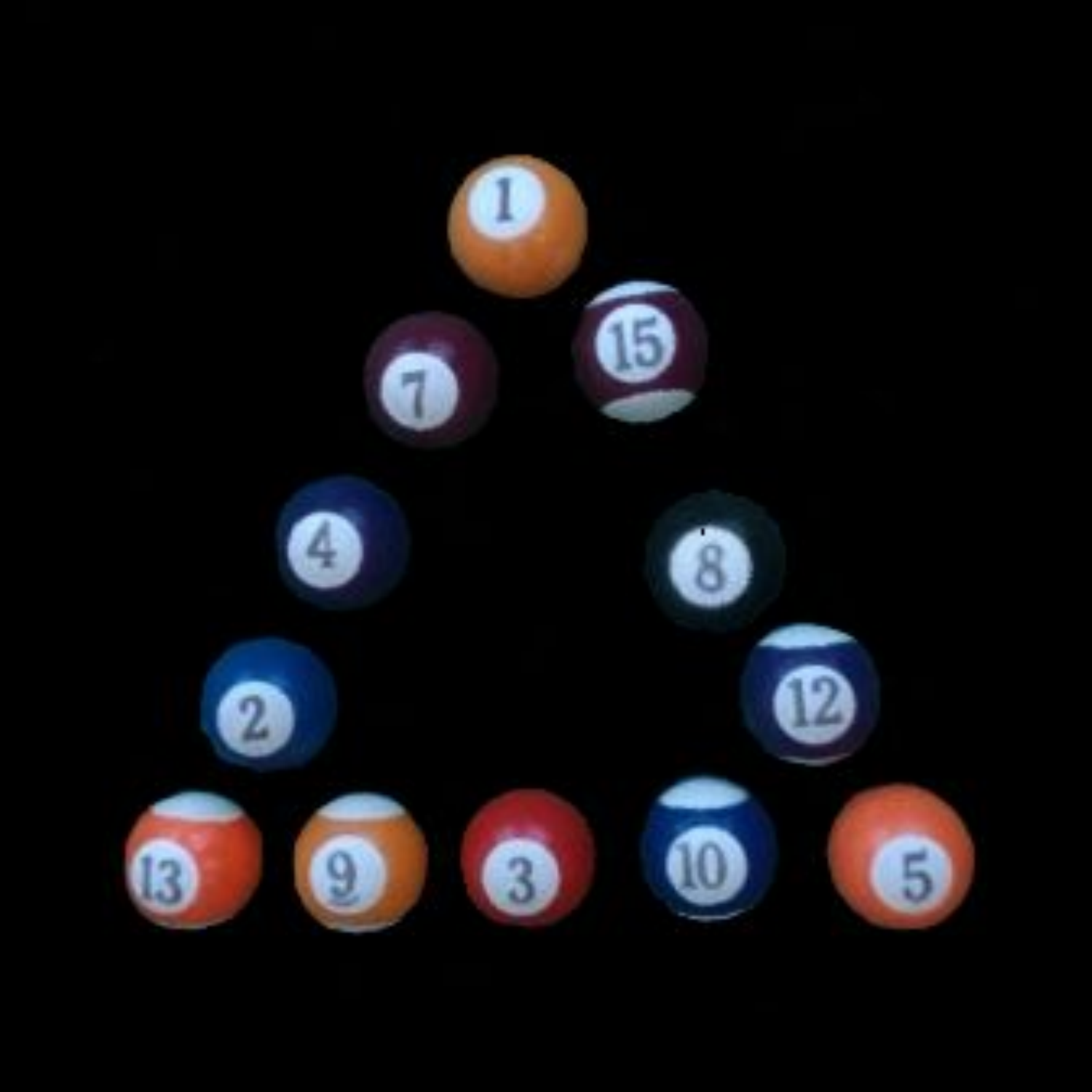}}
    \setcounter{subfigure}{0}\\\vspace{-0.22cm}
    \subfigure[Initial scene]{\includegraphics[width=0.3\linewidth]{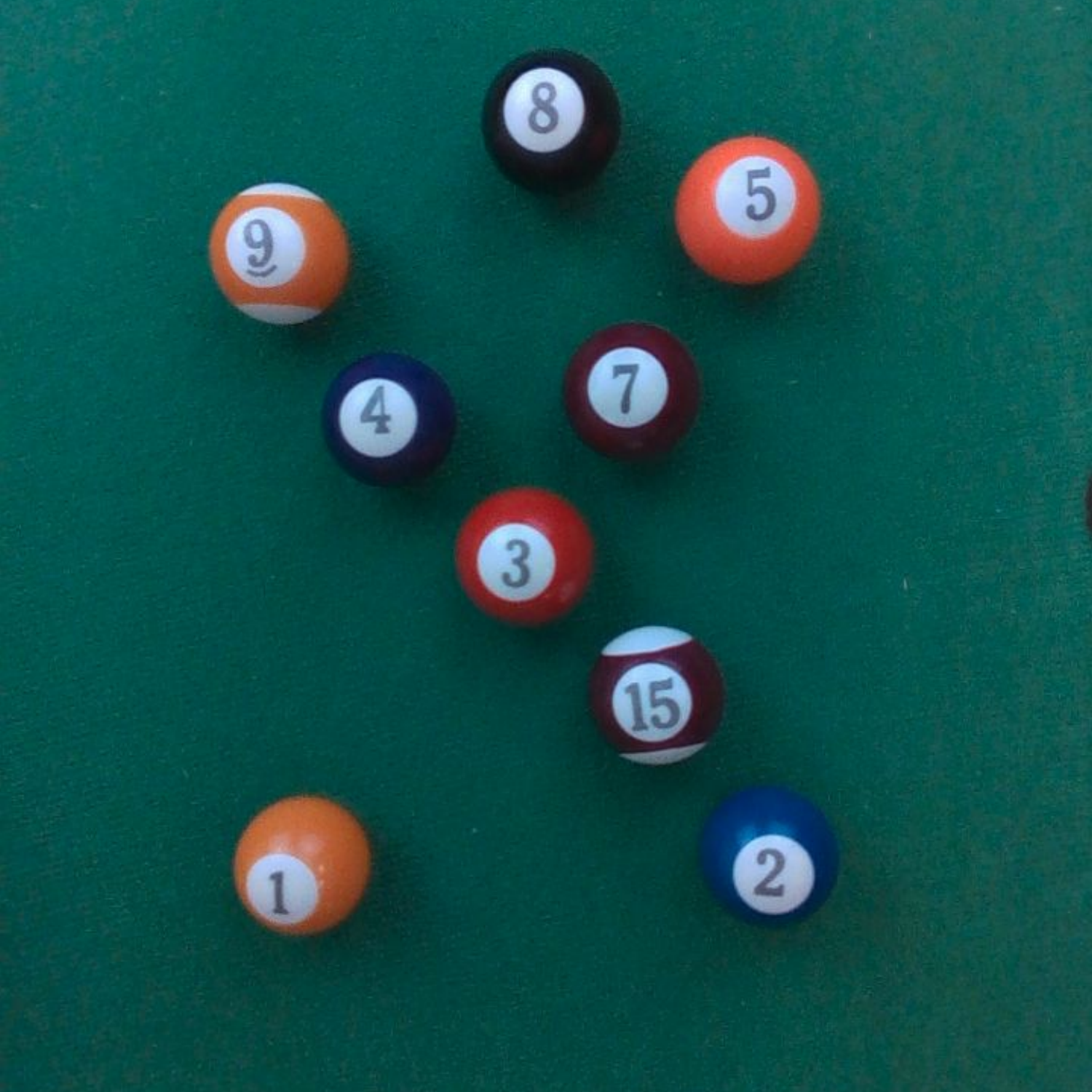}}
    \subfigure[Heatmap]{\includegraphics[width=0.3\linewidth]{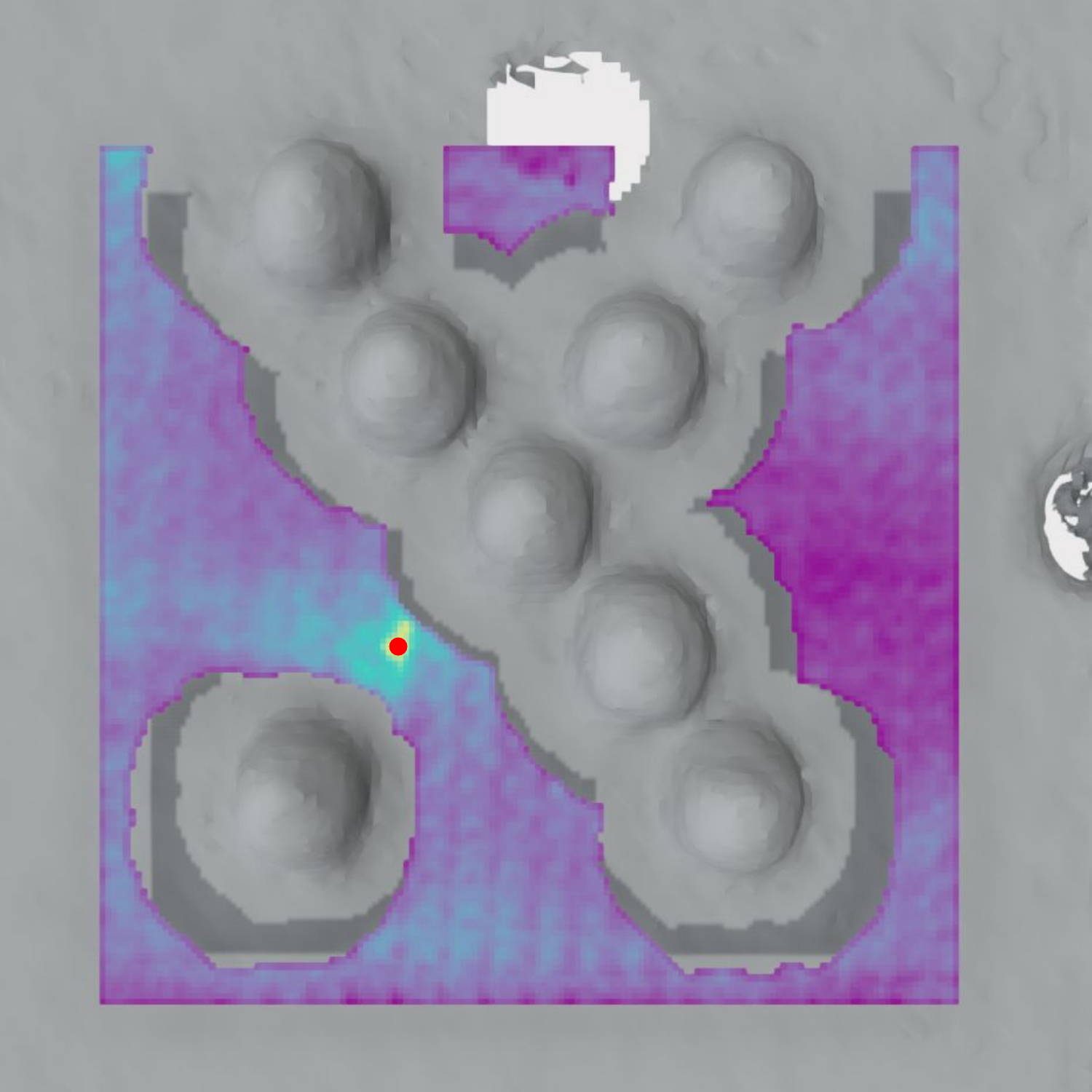}}
    \subfigure[Max score render]{\includegraphics[width=0.3\linewidth]{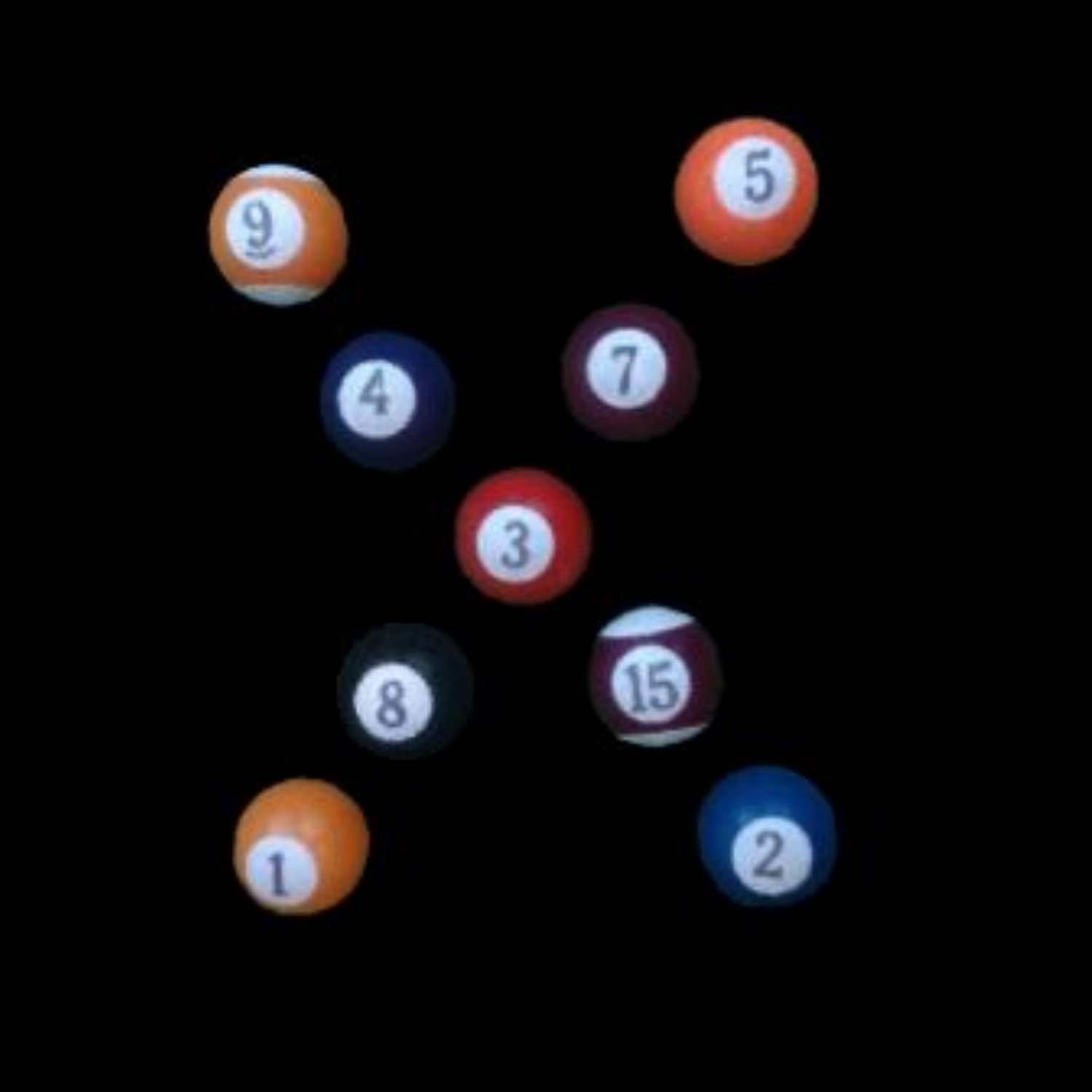}}
    \caption{Qualitative results from the pool ball scene for the tasks \textit{``in triangle''} (top row) and \textit{``in X shape''} (bottom row). The red dot is used to highlight the high-scoring area.}
    \label{fig: pool table scene}
    \vspace{-0.1cm}
\end{figure}

In this scene, we test our method on geometric relations involving many objects: the method must form a triangle out of 12 pool balls, and an X shape out of 9, by placing the final black ball in the correct position (in imagination). Positions are sampled at a 1mm resolution. In between the 5 roll-outs per method-task combination, we randomly take out a ball from the shape, and the method must complete the shape by placing the black ball. Results are shown in Fig \ref{fig: pool table scene}. The heatmap shows a high-scoring mode near the optimal pose for each task, suggesting that CLIP can understand geometric relations involving many objects. Success rates are in Table \ref{tab:results-pool}. DALL-E-Bot often fails due to the matching problem as before, which our evaluative approach avoids. \method{Physics-Only}'s low success rate shows that using CLIP for semantic guidance is useful. Interestingly, \method{D2R-Distract} performs well because there are no distractors here, and the green pool table background seems to provide helpful context to CLIP.

\begin{table}[btp]
    \centering
    \caption{Success rates for the pool ball scene (\%).}
    \label{tab:results-pool}
    \begin{tabular}{lcccc}
        \toprule
        Method & \textit{in X shape} & \textit{in triangle} & \textit{mean} \\
        \midrule
        D2R-Vis-Prior & 0 & 0 & 0 \\
        Physics-Only & 20 & 0 & 10 \\
        DALL-E-Bot \cite{dallebot} & 0 & 60 & 30 \\
        D2R-No-Norm & 80 & 40 & 60 \\
        D2R-One-View & 20 & \textbf{100} & 60 \\
        D2R-No-Smooth & 80 & 80 & 80 \\
        Dream2Real & \textbf{100} & 80 & 90 \\
        D2R-Distract & \textbf{100} & \textbf{100} & \textbf{100} \\
        \bottomrule
    \end{tabular}
    \vspace{-0.3cm}
\end{table}

\subsection{6-DoF Rearrangement in a 3D Scene}\label{ss:shelf}
\begin{figure}[tbp]
    \centering
    \subfigure{\label{fig:shelf_row_heatmap}\includegraphics[width=0.3\linewidth]{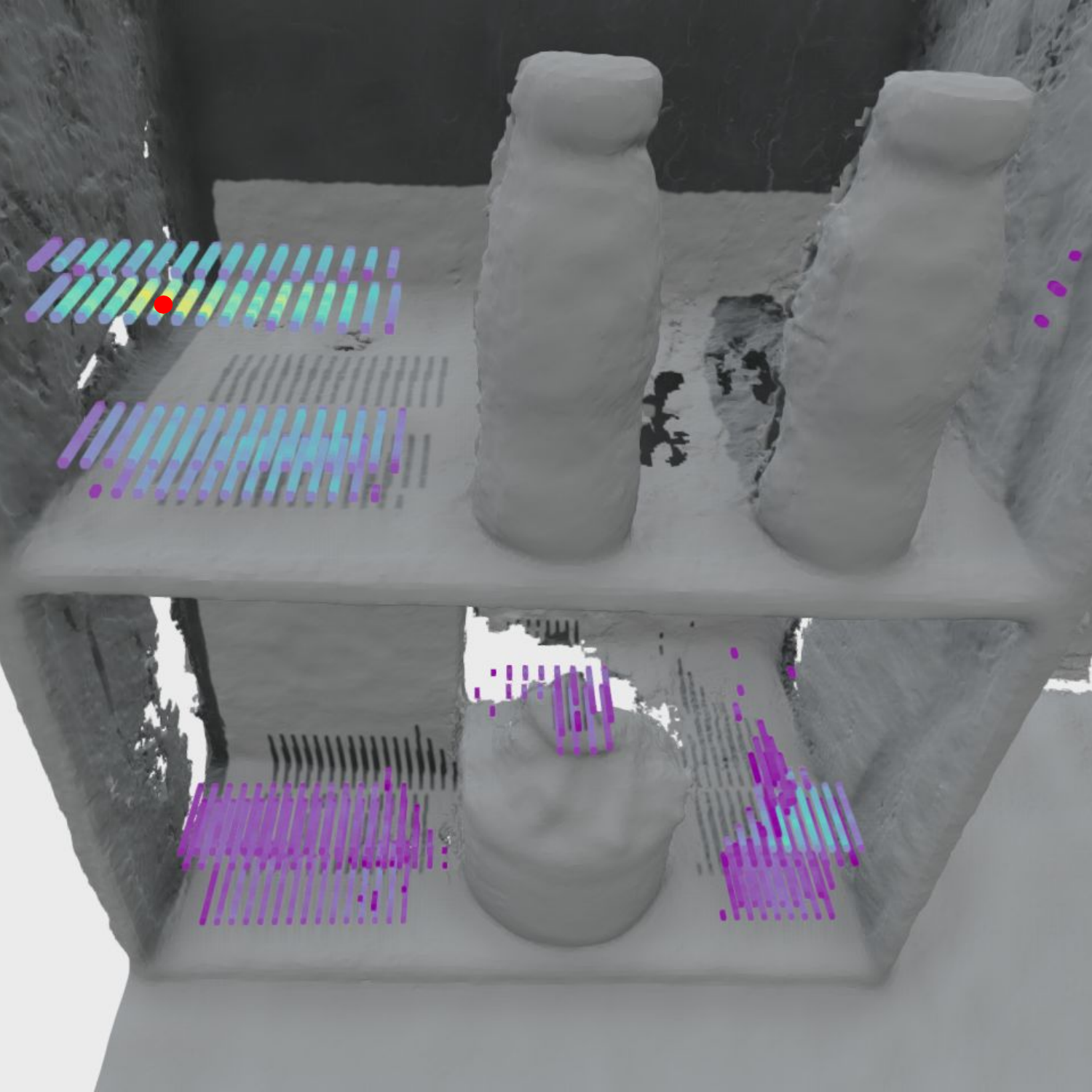}}
    \subfigure{\label{fig: shelf_book_heatmap}\includegraphics[width=0.3\linewidth]{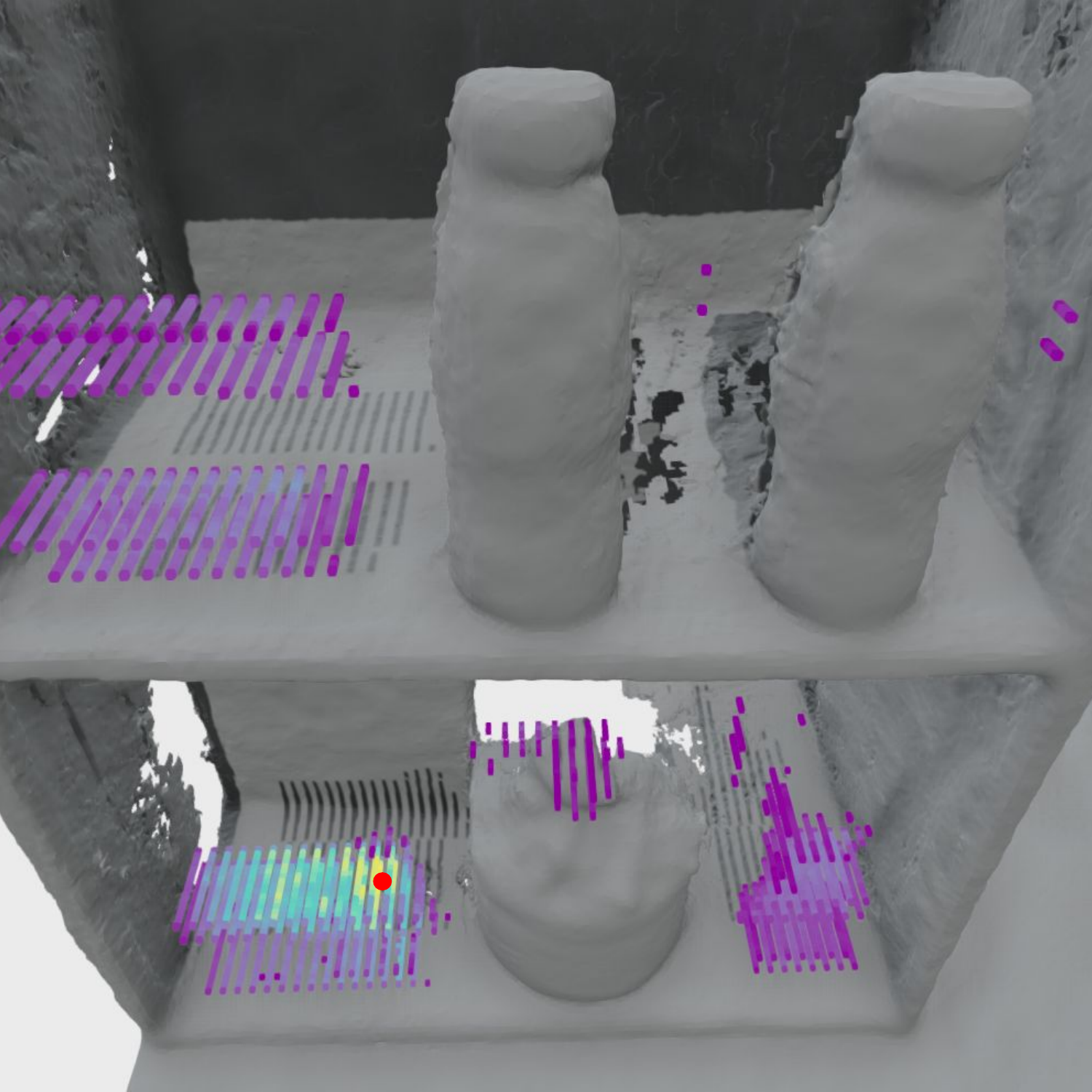}}
    \subfigure{\label{fig: shelf_plant_heatmap}\includegraphics[width=0.3\linewidth]{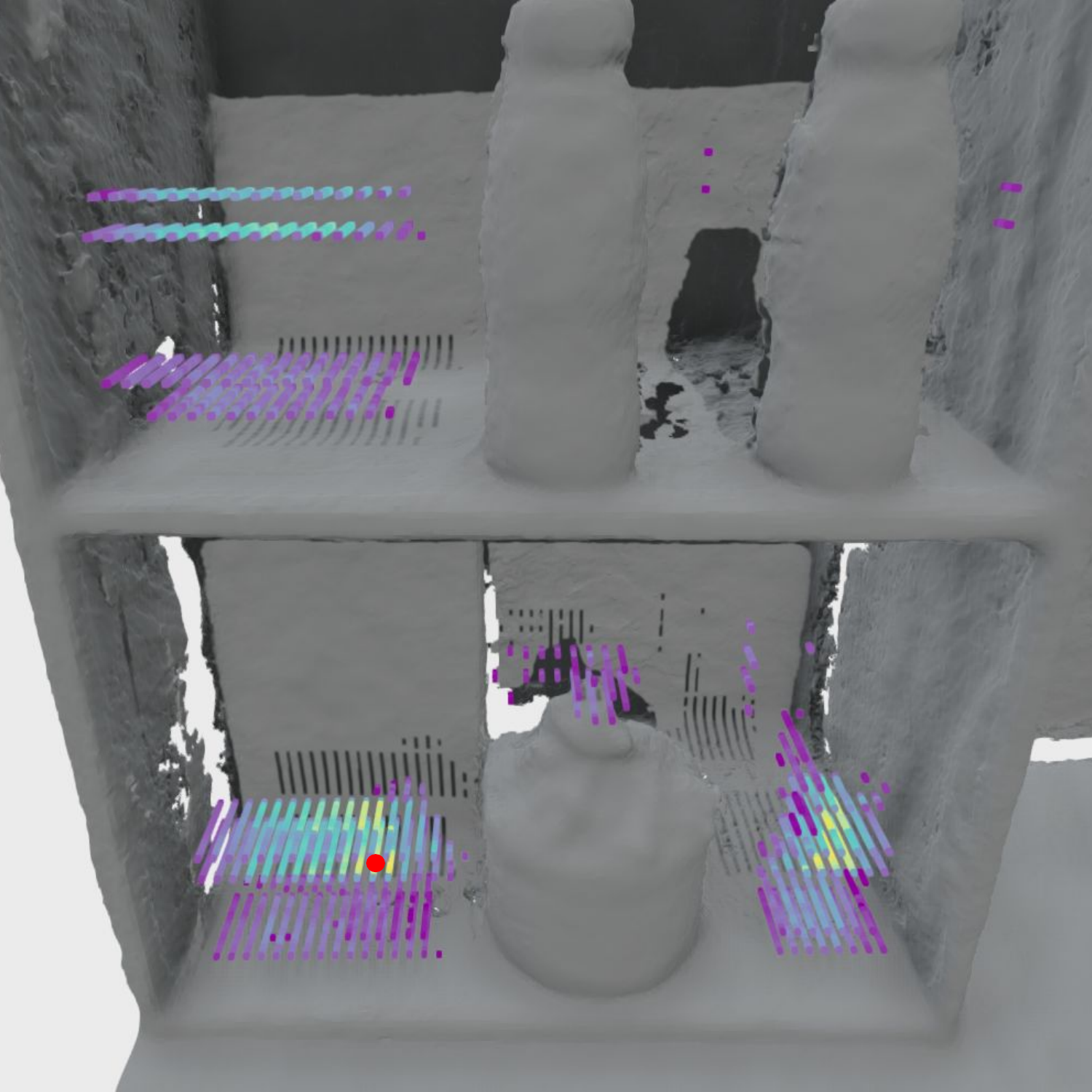}}
    \setcounter{subfigure}{0}\\\vspace{-0.22cm}
    \subfigure[\textit{``bottles in a row''}]{\label{fig:shelf_row_render}\includegraphics[width=0.3\linewidth]{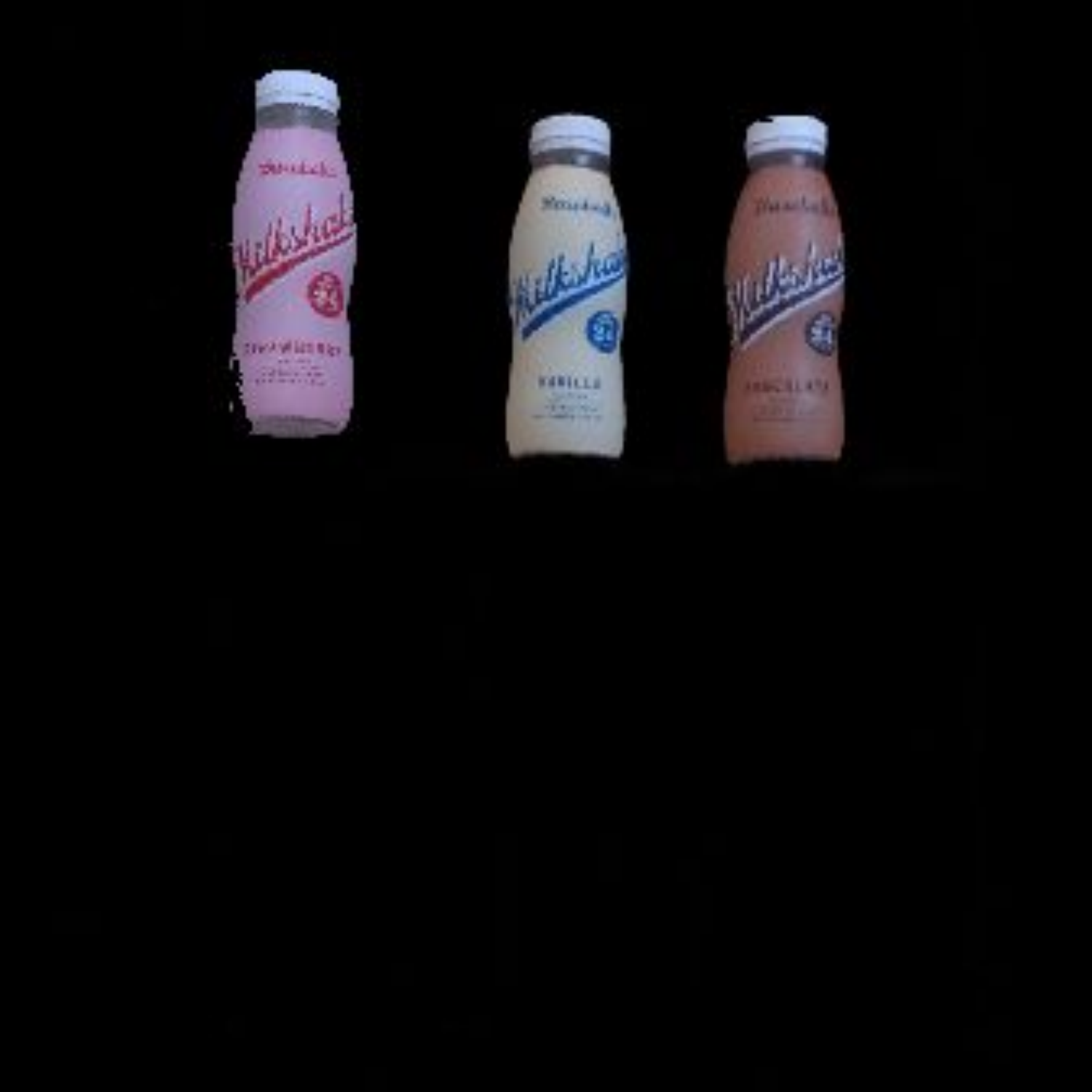}}
    \subfigure[\textit{``in front of book''}]{\label{fig:shelf_book_render}\includegraphics[width=0.3\linewidth]{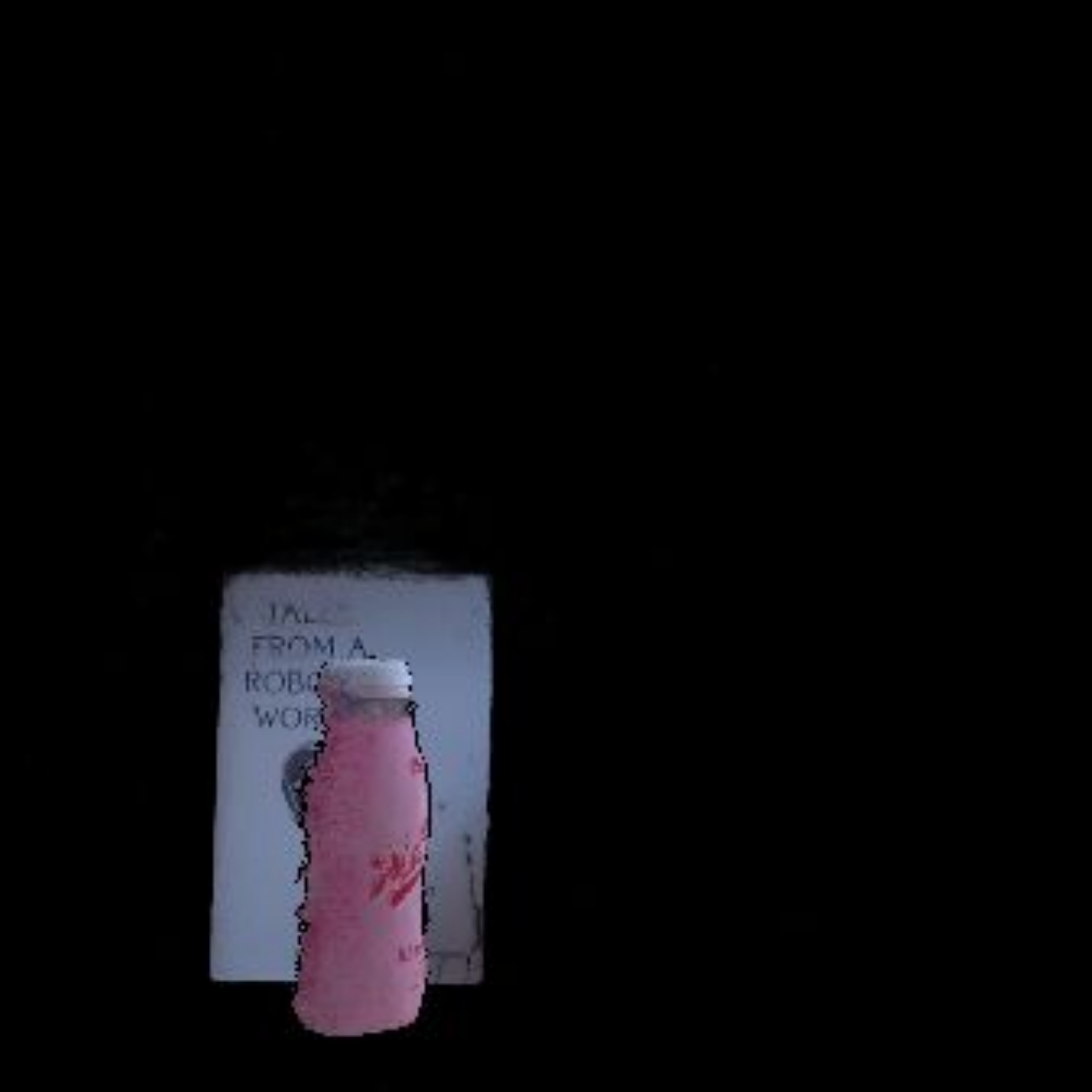}}
    \subfigure[\textit{``near plant''}]{\label{fig:shelf_plant_render}\includegraphics[width=0.3\linewidth]{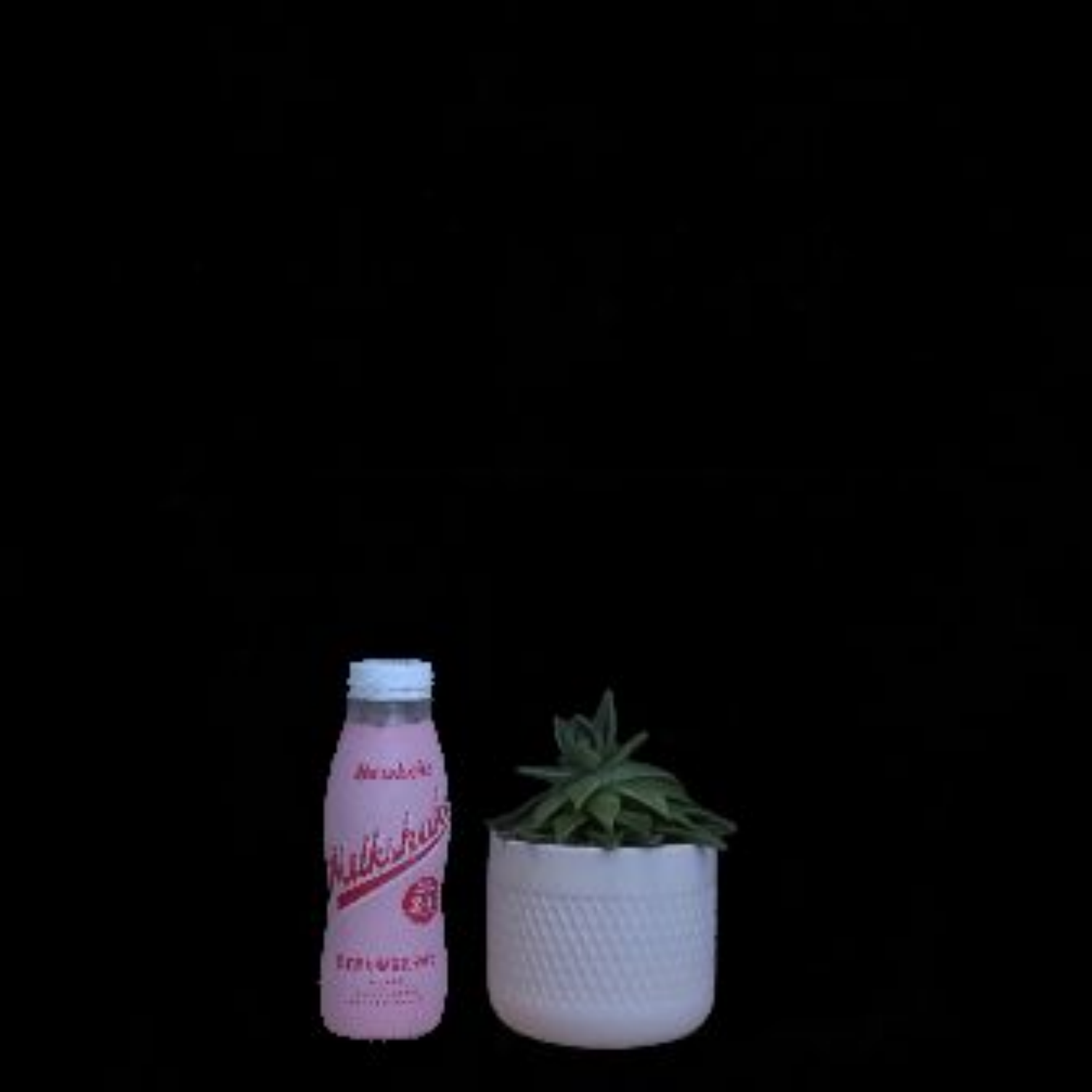}}
    \caption{Results for the three tasks on the shelf scene, with heatmaps (top row) and the highest-scoring renders (bottom).}
    \label{fig:qual-shelf}
\end{figure}

Here we test our method on a 3D shelf scene (see Figure \ref{fig:all-scenes}). Our method must perform 6-DoF rearrangement (in imagination) to pick up the bottle lying on the table and position it upright on the shelf. There are 3 tasks: making the bottles into a row, placing the bottle in front of the book, and placing the bottle near the plant. In this scene, we sample 24 orientations at each position (i.e. discretise coarsely into $\pi / 2$ orientations around each of the coordinate axes). In between each of the 10 roll-outs per method-task combination, we move and rotate the bottle around the table and shuffle some of the objects on the shelf. The heatmaps for each task are in Figure \ref{fig:qual-shelf}. We compare several interesting variations of our method in Figure \ref{fig:bar-chart}. \method{Physics-Only} rarely guesses the semantically correct upright orientation, showing that this is a challenging problem which our method addresses. Interestingly, the \method{D2R-One-View} baseline often fails due to incorrect object identification, whereas our approach which integrates captions across views is more robust. This shows that our multi-view approach is better suited to 6-DoF scenes.

\subsection{Demonstrating Physical Execution}\label{ss:exec-exps}

\begin{figure}[tbp]
    \centerline{\includegraphics[width=1.0\linewidth]{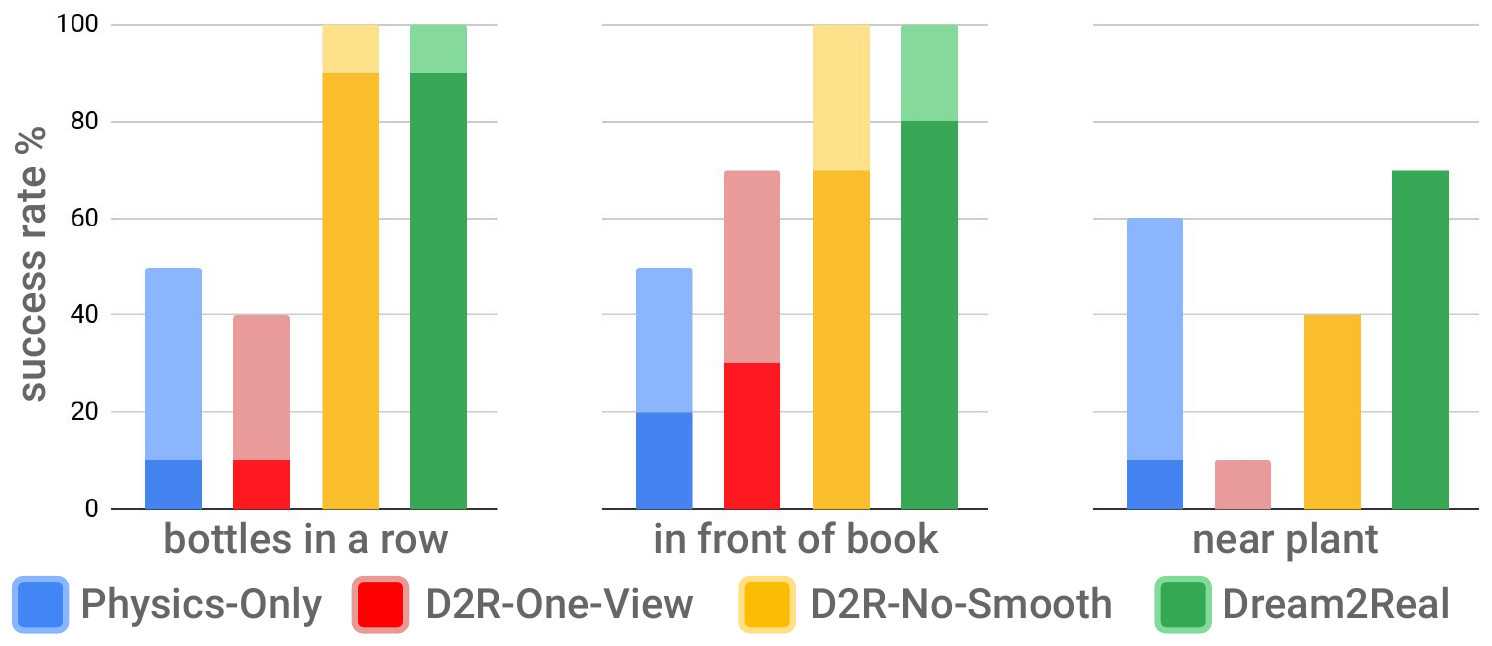}}
    \caption{Success rates for the shelf scene. A darker bar shows the success rate for predicting the full 6-DoF pose, and a lighter bar on top indicates roll-outs where the method correctly predicted the position but not the orientation.}
    \label{fig:bar-chart}
\end{figure}

\begin{figure}[tbp]
    \centerline{\includegraphics[width=0.65\linewidth]{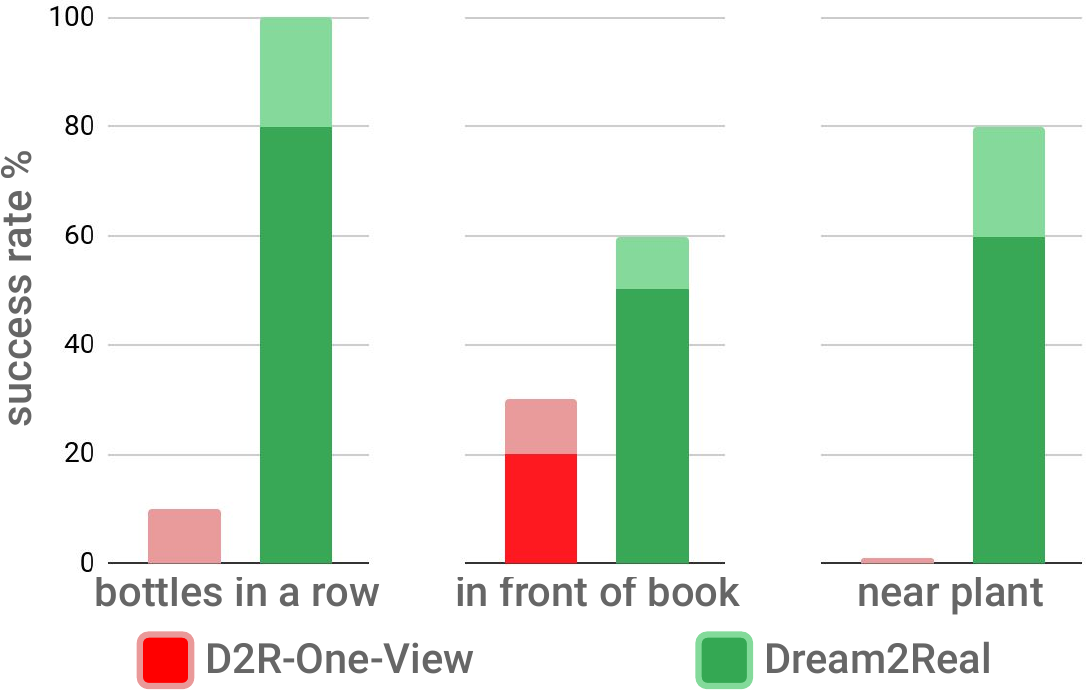}}
    \caption{Success rates for robotic execution. A darker bar shows the success rate for placing the object, and a lighter bar on top indicates roll-outs where the method correctly predicted the 6-DoF pose but did not execute successfully.}
    \label{fig:robot-bar-chart}
    \vspace{-0.2cm}
\end{figure}

Although the main contribution of our paper is predicting the goal pose, here we also demonstrate how a robot can pick and place objects into those goal poses. Robot videos for all scenes are available on our website: \textcolor{blue}{\href{https://www.robot-learning.uk/dream2real}{https://www.robot-learning.uk/dream2real}}. In Figure \ref{fig:robot-bar-chart}, we compare our multi-view method with \method{D2R-One-View} on the 6-DoF tasks in the shelf scene with robotic execution. Since we now automatically eliminate unreachable poses, results will differ from Figure \ref{fig:bar-chart}. We find that the single-view baseline fails more often due to incomplete reconstruction, which impacts both goal pose prediction and collision-free motion planning. This shows that our multi-view Dream2Real framework can be used to perform 6-DoF rearrangement on a real robot.


\section{Discussion}

\textbf{Limitations and future work}. \textbf{(1)} Low-tolerance tasks like insertion would require sampling poses more densely, which is computationally expensive. \textbf{(2)} Running time is a limitation of the current implementation of our framework. Scanning the scene (e.g. when the robot first observes a new room) takes 3-5 minutes. This can be reduced in future work using sparse NeRFs \cite{pixelnerf} or generative image-to-3D methods \cite{realfusion,zero1to3}. After the user instruction is received, it currently takes approximately 6 minutes to render, check and score all the poses. In future work this can be sped up with an iterative coarse-to-fine approach: sampling poses sparsely at first, and then more densely in the higher-scoring regions. \textbf{(3)} As shown in prior research \cite{negclip}, CLIP can exhibit bag-of-words behaviour, i.e. CLIP performs poorly on goal captions where the order of words matters. E.g. \textit{``a fork to the left of a knife''} often places the knife to the left of the fork instead. However, as shown empirically, CLIP performs well on several useful tasks and even complex spatial relations. As VLMs improve in the future, this limitation will be less significant. To further improve the robustness of CLIP scoring, in future work arrangements could be rendered from multiple views, and the CLIP scores aggregated from different perspectives.

\textbf{Conclusions}. We show for the first time how 2D VLMs can be used to perform language-conditioned 3D object rearrangement zero-shot, without needing to collect any example arrangements. Encouraging results show that our method Dream2Real can complete everyday rearrangement tasks, understands complex multi-object relations, and is robust to distractors. While prior work \cite{dallebot} uses VLMs to \textit{generate} goal images, our approach uses VLMs to \textit{evaluate} candidate images. This improves performance in tabletop experiments and unlocks the use of 2D VLMs for 3D tasks.

\clearpage
\section*{Acknowledgements}

We thank our colleagues from the Robot Learning Lab and the Dyson Robotics Lab for helpful discussions. In particular, we would like to thank Andrew J. Davison, Eric Dexheimer, Xin Kong, Hidenobu Matsuki, Marwan Taher, Vitalis Vosylius, and Kentaro Wada. In addition, we are grateful to Shikun Liu for lending his expertise in diagram design. This research was supported by: Dyson Technology Ltd, EPSRC Prosperity Partnerships (EP/S036636/1), and the Royal Academy of Engineering under the Research Fellowship Scheme. For the purpose of open access, the authors have applied a Creative Commons Attribution (CC BY) license to any Accepted Manuscript version arising.


\bibliographystyle{IEEEtran}
\bibliography{dream2real}

\end{document}